%% file: arxiv.tex
\documentclass{article}

\usepackage[preprint]{corl_2023} 
\input{packages}
\title{\ours: Dialectic Multi-\underline{Ro}bot \underline{Co}llaboration \\ with Large Language Models}
\author{
  Zhao Mandi\\
  Columbia University\\ 
  \And
  Shreeya Jain \\
  Columbia University\\ 
  \And
  Shuran Song \\
  Columbia University\\ 
}
\begin{document}
\maketitle

\begin{abstract}
We propose a novel approach to multi-robot collaboration that harnesses the power of pre-trained large language models (LLMs) for both high-level communication and low-level path planning. Robots are equipped with LLMs to discuss and collectively reason task strategies. They then generate sub-task plans and task space waypoint paths, which are used by a multi-arm motion planner to accelerate trajectory planning. We also provide feedback from the environment, such as collision checking, and prompt the LLM agents to improve their plan and waypoints in-context. For evaluation, we introduce \benchmark, a 6-task benchmark covering a wide range of multi-robot collaboration scenarios, accompanied by a text-only dataset for agent representation and reasoning. We experimentally demonstrate the effectiveness of our approach -- it achieves high success rates across all tasks in \benchmark~and adapts to variations in task semantics. Our dialog setup offers high interpretability and flexibility -- in real world experiments, we show \ours~easily incorporates human-in-the-loop, where a user can communicate and collaborate with a robot agent to complete tasks together. See project website \href{https://project-roco.github.io/}{project-roco.github.io} for videos and code.
\end{abstract}



\input{text/introduction}

\input{text/statement}
\input{text/method}

\input{text/benchmark}
\label{sec:result} 
\input{text/result}

\input{text/reasoning-dataset}
\vspace{-2mm}
\section{Limitation}
\vspace{-3mm}
\textbf{Oracle state information in simulation.} \ours~assumes perception (e.g., object detection, pose estimation and collision-checking) is accurate. This assumption makes our method prone to failure in cases where perfect perception is not available: this is reflected in our real-world experiments, where the pre-trained object detection produces errors that can cause planning mistakes. 

\textbf{Open-loop execution.} The motion trajectories from our planner are executed by robots in an open-loop fashion and lead to potential errors. Due to the layer of abstraction in scene and action descriptions, LLMs can't recognize or find means to handle such execution-level errors.

\textbf{LLM-query Efficiency.} We rely on querying pre-trained LLMs for generating every single response in an agent's dialog, which can be cost expensive and dependant on the LLM's reaction time. Response delay from LLM querying is not desirable for tasks that are dynamic or speed sensitive. 
\input{text/relatedwork}

\vspace{-3mm}
\section{Conclusion}
\vspace{-3mm}
\label{sec:conclusion} 
We present \ours, a new framework for multi-robot collaboration that leverages large language models (LLMs) for robot coordination and planning. We introduce \benchmark, a 6-task benchmark for multi-robot manipulation to be open-sourced to the broader research community. We empirically demonstrate the generality of our approach and many desirable properties such as few-shot adaptation to varying task semantics, while identifying limitations and room for improvement. Our work falls in line with recent literature that explores harnessing the power of LLMs for robotic applications, and points to many exciting opportunities for future research in this direction.

\acknowledgments{ 
This work was supported in part by NSF Award \#2143601, \#2037101, and \#2132519. We would like to thank Google for the UR5 robot hardware.  The views and conclusions contained herein are those of the authors and should not be interpreted as necessarily representing the official policies, either expressed or implied, of the sponsors. The authors would like to thank Zeyi Liu, Zhenjia Xu, Huy Ha, Cheng Chi, Samir Gadre, Mengda Xu, and Dominik Bauer for their fruitful discussions throughout the project and for providing helpful feedback on initial drafts of the manuscript.}
\bibliography{reference}  

\input{text/appendix}
\end{document}

%% file: packages.tex
\usepackage{graphicx} 
\usepackage{wrapfig}
\usepackage{lipsum}
\usepackage{enumitem}
\usepackage{booktabs}
\usepackage{pifont}
\usepackage{soul}
\usepackage{amsmath}
\usepackage{amssymb}
\usepackage{multirow} 
\usepackage{array}  
\usepackage{multicol} 
\usepackage{times} 
\usepackage{siunitx}
\usepackage{gensymb} 
\usepackage{pifont}
\usepackage{soul}
\usepackage[symbol]{footmisc}

\usepackage{mdframed}

\usepackage{algorithm}
\usepackage{algpseudocode} 
\usepackage{todonotes}
\usepackage{subcaption}
\usepackage{xcolor}
\usepackage{adjustbox}
\usepackage{pdflscape}
\usepackage{bbm}
\usepackage{hyperref}
\hypersetup{
    colorlinks=true,
    linkcolor=blue,
    urlcolor=blue,
}
\newcommand{\ours}{RoCo}
\newcommand{\benchmark}{RoCoBench}
 
\renewcommand\fbox{\fcolorbox{light-gray}{bg-gray}}
\newcommand{\promptblock}[1]{\vspace{1pt}\noindent\fbox{\parbox{0.98\linewidth}{{\textcolor{black}{{#1}}}}}\vspace{1pt}}

\newcommand{\commenttext}[1]{\textcolor{light-gray}{#1}}
\newcommand{\llmcompletion}[1]{\colorbox{highlight}{#1}}
\newcommand{\greencompletion}[1]{\colorbox{light-green}{#1}}
\newcommand{\yellowcompletion}[1]{\colorbox{light-yellow}{#1}}

\newcommand{\blank}[1]{\textcolor{brown}{#1}}
\definecolor{dark-gray}{HTML}{A9A9A9} %
\definecolor{light-gray}{HTML}{b7b7b7} %
\definecolor{light-green}{HTML}{dcf8c6}
\definecolor{light-blue}{HTML}{a5cdfb}
\definecolor{bg-gray}{HTML}{F8F8F8} 
	
\definecolor{dark-green}{HTML}{6aa84f} 
\definecolor{highlight}{HTML}{cfe2f3} 
\definecolor{blue}{HTML}{2B60DE} 
\definecolor{brown}{HTML}{9F8C76}
\definecolor{pink}{HTML}{D88782}
\definecolor{dark-yellow}{HTML}{F6AE2D}
\definecolor{light-yellow}{HTML}{ebdc7d}

%% file: text/introduction.tex
\begin{figure}[h!]
     \centering \includegraphics[width=0.9\linewidth]{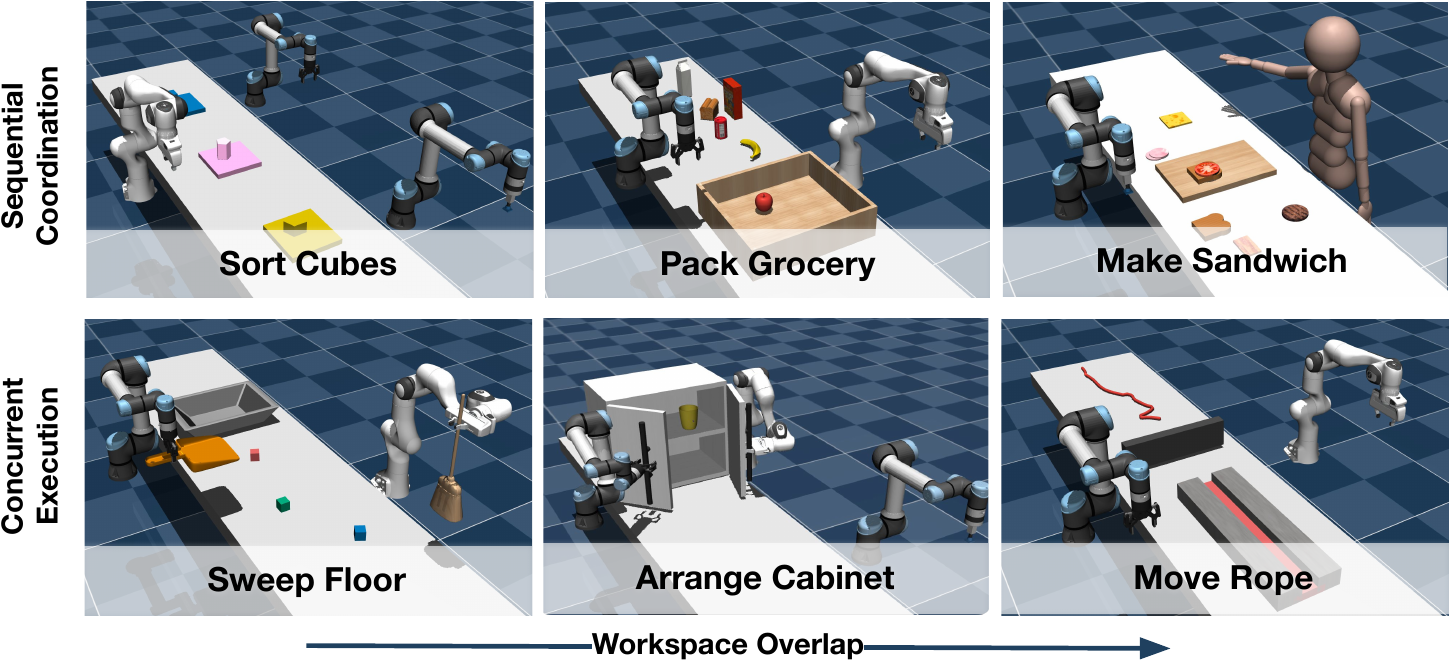}
     \caption{\footnotesize We propose \ours, a unified approach for multi-robot collaboration that leverages LLMs for both high-level task coordination and low-level motion planning. We demonstrate its utility on \benchmark, a benchmark we introduce that includes a diverse set of challenges in collaboration task scenarios.}
     \vspace{-5mm}
     \label{fig:task}
 \end{figure}
 
 \section{Introduction}
 \vspace{-2mm}
Multi-robot systems are intriguing for their promise of enhancing task productivity, but are faced with various challenges. For robots to effectively split and allocate the work, it requires high-level understanding of a task, and consideration of each robot's capabilities such as reach range or payload. Another challenge lies in low-level motion planning: as the configuration space grows with the number of robots, finding collision-free motion plans becomes exponentially difficult. Finally, traditional multi-robot systems typically require task-specific engineering, hence compromise generalization: with much of the task structures pre-defined, these systems are incapable of adapting to new scenarios or variations in a task. 

We propose \textbf{\ours}, a zero-shot multi-robot collaboration method to address the above challenges. Our approach includes three key components: 
\begin{itemize}[leftmargin=*]
    \item \vspace{-2mm} \textbf{Dialogue-style task-coordination}: To facilitate information exchange and task reasoning, we let robots `talk' among themselves by delegating each robot to an LLM agent in a dialog, which allows robots to discuss the task in natural language, with high interpretability for supervision.  
    \item \textbf{Feedback-improved Sub-task Plan Generated by LLMs}: The multi-agent dialog ends with a sub-task plan for each agent (e.g. pick up object). We provide a set of environment validations and feedback (e.g. IK failures or collision) to the LLM agents until a valid plan is proposed. 
    \item \textbf{LLM-informed Motion-Planning in Joint Space}: From the validated sub-task plan, we extract goal configurations in the robots' joint space, and use a centralized RRT-sampler to plan motion trajectories. We explore a less-studied capability of LLM: 3D spatial reasoning. Given the start, goal, and obstacle locations in task space, we show LLMs can generate waypoint paths that incorporate high-level task semantics and environmental constraints, and significantly reduce the motion planner's sample complexity.  
\end{itemize}  
\vspace{-2mm}
We next introduce \benchmark, a benchmark with 6 multi-robot manipulation tasks. We experimentally demonstrate the effectiveness of \ours~on the benchmark tasks: by leveraging the commonsense knowledge captured by large language models (LLMs), \ours~is flexible in handling a variety of collaboration scenarios without any task-specific training. 
 
In summary, we propose a novel approach to multi-robot collaboration, supported by two technical contributions:
1) An LLM-based multi-robot framework (\textbf{\ours}) that is flexible in handling a large variety of tasks with improved task-level coordination and action-level motion planning; 2) A new benchmark (\textbf{\benchmark}) for multi-robot manipulation to systematically evaluate these capabilities. It includes a suite of tasks that are designed to examine the flexibility and generality of the algorithm in handling different task semantics (e.g., sequential or concurrent), different levels of workspace overlaps, and varying agent capabilities (e.g., reach range and end-effector types) and embodiments (e.g., 6DoF UR5, 7DoF Franka, 20DoF Humanoid).

%% file: text/statement.tex
\vspace{-1mm}
\section{Preliminaries}
\vspace{-3mm}
\textbf{Task Assumption.}
We consider a cooperative multi-agent task environment with $N$ robots, a finite time horizon $T$, full observation space $O$. Each agent $n$ has observation space $\Omega^n \subset O$. Agents may have asymmetric observation spaces and capabilities, which stresses the need for communication. We manually define description functions $f$ that translate task semantics and observations at a time-step $t$ into natural language prompts: $l^n_t = f^n(o_t), o_t \in \Omega^{n}$. We also define parsing functions that map LLM outputs (e.g. text string ``PICK object") to the corresponding sub-task, which can be described by one or more gripper goal configurations.


\textbf{Multi-arm Path Planning.}
Given a sub-task, we use centralized motion planning that finds trajectories for all robots to reach their goal configurations. Let $\mathcal{X} \in \mathbb{R}^{d}$ denote the joint configuration space of all $N$ robot arms 
and $\mathcal{X}_{ob}$ be the obstacle region in the configuration space, then collision-free space $\mathcal{X}_{free} = \mathcal{X}\backslash\mathcal{X}_{ob}$. Given an initial condition $x_{init} \in \mathcal{X}_{free}$, a goal region
$\mathcal{X}_{goal} \in \mathcal{X}_{free}$, the motion planner finds an optimal $\sigma^*:[0,1]\rightarrow \mathcal{X}$ that satisfies: $\sigma^*(0) = x_{init}$, $\sigma^*(1) \in x_{goal}$. This resulting $\sigma^*$ is used by the robots' arm joint controllers to execute in open-loop.  

%% file: text/method.tex
\vspace{-1mm}
\section{Multi-Robot Collaboration with LLMs}
\vspace{-3mm}
 \begin{figure}[tp!]
     \centering \includegraphics[width=0.99\linewidth]{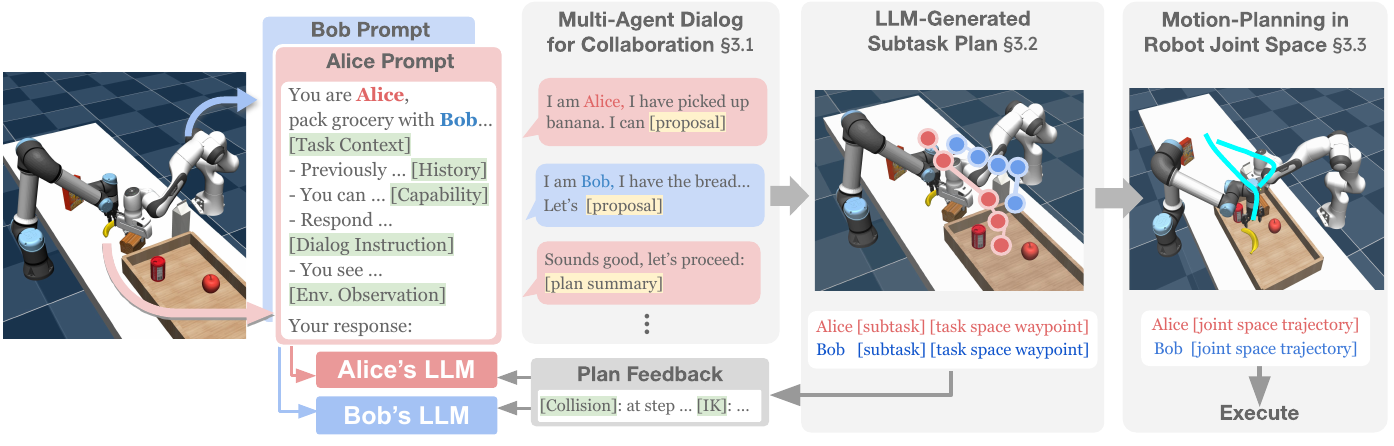}
     \caption{\footnotesize
     \ours~consists of three main components: 1) \textbf{Multi-agent dialog via LLMs}: each robot is equipped with an LLM that `talks' on its behalf, enabling a discussion of task strategy. 2) \textbf{LLM-Generated Sub-task Plan}: the dialog ends with a proposal of sub-task plan, including optionally a path of task space waypoints, and environment feedback on invalid plans are provided for the agents to improve. 3) \textbf{Multi-arm motion planning}: A validated sub-task plan then produces goal configurations for the robot arms, which are used by a centralized multi-arm motion planner that outputs trajectories for each robot.}
     \vspace{-3mm}
     \label{fig:method}
 \end{figure}
We present \ours, a novel method for multi-robot collaboration that leverages LLMs for robot communication and motion planning. The three key components in our method are demonstrated in Fig. \ref{fig:method} and described below:
\vspace{-3mm}
\subsection{Multi-Agent Dialog via LLMs}
\vspace{-2mm}
We assume multi-agent task environments with asymmetric observation space and skill capabilities, which means agents can't coordinate meaningfully without first communicating with each other. We leverage pre-trained LLMs to facilitate this communication. Specifically, before each environment interaction, we set up one round of dialog where each robot is delegated an LLM-generated agent, which receives information that is unique to this robot and must respond strictly according to its role (e.g. ``I am Alice, I can reach ...").

For each agent's LLM prompt, we use a shared overall structure but with agent-specific content, varied with each robot's individual status. The prompt is composed of the following key components:
\begin{enumerate}[leftmargin=*]
\footnotesize
\vspace{-1mm}
\item \vspace{-1mm} \textbf{Task Context}: describes overall objective of the task.
\item \vspace{-1mm} \textbf{Round History}: past dialog and executed actions from previous rounds.
\item \vspace{-1mm} \textbf{Agent Capability}: the agent's available skills and constraints.
\item \vspace{-1mm} \textbf{Communication Instruction}: how to respond to other agents and properly format outputs.
\item \vspace{-1mm} \textbf{Current Observation}: unique to each agent's status, plus previous responses in the current dialog.
\item \vspace{-1mm} \textbf{Plan Feedback}: (optional) reasons why a previous sub-task plan failed.
\end{enumerate}
\vspace{-2mm}
We use the following communication protocol to move the dialog forward: each agent is asked (instructions are given in Communication Instruction part of the prompt) to end their response by deciding between two options: 1) indicate others to proceed the discussion; 2) summarize everyone's actions and make a final action proposal; this is allowed only if each agent has responded at least once in the current dialog round. This protocol is designed to allow the agents to freely discuss, while guaranteeing one sub-task plan will be proposed within a finite number of exchanges. 


\vspace{-1mm}
\subsection{LLM-Generated Sub-task Plan} 
\vspace{-2mm}

\begin{wrapfigure}{r}{0.22\textwidth} 
\centering
\vspace{-15mm}
\includegraphics[width=0.2\textwidth]{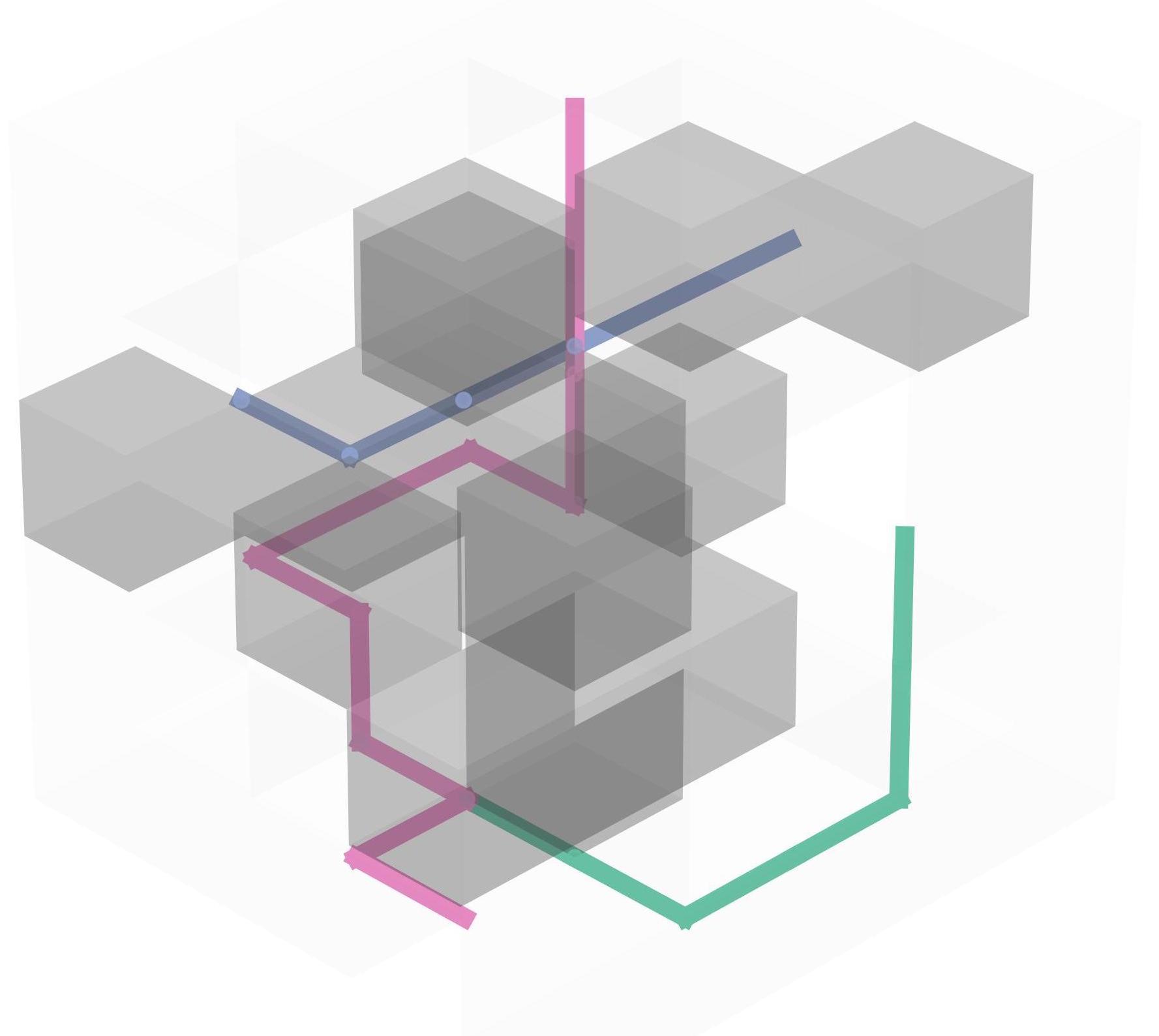} 
\vspace{-2mm}
\caption{\footnotesize Spatial reasoning by LLM.  GPT-4 plans 3D collision-free paths for three agents.} \vspace{-6mm}
\label{fig:toy_grid} 
\end{wrapfigure} 
Once a round of dialog ends, the last speaking agent summarizes a `sub-task plan', where each agent gets one sub-task  (e.g. pick up object) and optionally a path of 3D waypoints in the task space. This sub-task plan is first passed through a set of validations before going into execution. If any of the checks fail, the feedback is appended to each agent's prompt and another round of dialog begins. The validations are conducted in the following order, and each one assumes the previous check has passed (e.g. a plan must be parsed before checking for task and agent constraints):
\begin{enumerate}[leftmargin=*]
\vspace{-1mm}
\item \vspace{-1mm} \textbf{Text Parsing} ensures the plan follows desired format and contains all required keywords
\item \vspace{-1mm} \textbf{Task Constraints} checks whether each action complies with the task and agent constraints
\item \vspace{-1mm} \textbf{IK} checks whether each robot arm's target pose is feasible via inverse kinematics 
\item \vspace{-1mm} \textbf{Collision Checking} checks if the IK-solved arm join configurations cause collision
\item \vspace{-1mm} \textbf{Valid Waypoints} optionally, if a task requires path planning, each intermediate waypoint must be IK-solvable and collision-free, and all steps should be evenly spaced
\end{enumerate}
\vspace{-2mm} 
The agents are allowed to re-plan until reaching a maximum number of attempts, after which the current round ends without any execution and the next round begins. The episode is considered failed if the task is not completed within a finite number of rounds. 
\subsection{LLM-informed Motion Planning in Joint Space} 
\vspace{-2mm}
Once a sub-task plan passes all validations, we combine it with IK to produce a goal configuration jointly over all robot arms, and optionally, each step of the task space waypoint paths produces an intermediate goal configuration. The goal configuration(s) are passed to an RRT-based multi-arm motion planner that jointly plans across all robot arms, and outputs motion trajectories for each robot to execute in the environment, then the task moves on to the next round.

\textbf{Toy Example: 3D Spatial Reasoning Capability in LLMs.}
As a motivating example, we ask an LLM (GPT-4) to plan multi-agent paths in a 3D grid. We randomly sample 3 agents' (start, goal) coordinates and a set of obstacles, and prompt GPT-4 to plan collision-free paths. Given feedback on failed plans, allowing up to 5 attempts, GPT-4 obtains $86.7\%$ success rate over 30 runs, using on average $2.73$ attempts. See Fig.~\ref{fig:toy_grid} for an example output. See Appendix \ref{appendix:toy-problem} for further details and results. While this capability is encouraging, we found it to also be limited as the grid size and number of obstacles increase. 


%% file: text/benchmark.tex
\vspace{-3mm}
\section{Benchmark}
\vspace{-3mm}
\label{sec:benchmark}
\benchmark~is a suite of 6 multi-robot collaboration tasks in a tabletop manipulation setting. The tasks involve common-sense objects that are semantically easy to understand for LLMs, and span a repertoire of collaboration scenarios that require different robot communication and coordination behaviors.  See Appendix \ref{appendix:benchmark} for detailed documentation on the benchmark.
We remark three key properties that define each task, summarized in Table \ref{task-design-table}: 

\begin{enumerate}[leftmargin=*]
    \item \vspace{-1.2mm} Task decomposition: whether a task can be decomposed into sub-parts that can be completed in parallel or in certain order. Three tasks in \benchmark~have a sequential nature (e.g. Make Sandwich task requires food items to be stacked in correct order), while the other three tasks can be executed in parallel (e.g. objects in Pack Grocery task can be put into bin in any order).
    \item \vspace{-1.2mm} Observation space: how much of the task and environment information each robot agent receives. Three tasks provide shared observation of the task workspace, while the other three have a more asymmetric setup and robots must inquire each other to exchange knowledge. 
    \item \vspace{-1.2mm} Workspace overlap: proximity between operating robots; we rank each task from low, medium or high, where higher overlap  calls for more careful low-level coordination (e.g. Move Rope task requires manipulating the same object together).
\end{enumerate}   
\input{text/table-benchmark-design.tex}

%% file: text/table-benchmark-design.tex
\begin{wraptable}{r}{0.5\linewidth}
\centering
\vspace{-11mm}
\setlength{\tabcolsep}{0.06cm}
\centering
\scriptsize 
\begin{tabular}{@{}ccccccc@{}}
\toprule
& \begin{tabular}[c]{@{}c@{}}Sweep\\ Floor\end{tabular} 
& \begin{tabular}[c]{@{}c@{}}Pack \\ Grocery\end{tabular} 
& \begin{tabular}[c]{@{}c@{}}Move\\ Rope\end{tabular} 
& \begin{tabular}[c]{@{}c@{}}Arrange \\ Cabinet\end{tabular} 
& \begin{tabular}[c]{@{}c@{}}Make \\ Sandwich\end{tabular} 
& \begin{tabular}[c]{@{}c@{}}Sort\\ Cubes\end{tabular}  \\ \midrule
Task & Parallel & Parallel & Parallel  & Seq  & Seq & Seq          \\ \midrule
Observation  & Asym.    & Shared    & Shared    & Asym.      & Asym.     & Shared     \\ \midrule
Workspace  & Med     & Med    & High    & High        & Low        & Low           \\ \bottomrule  
\end{tabular}
\vspace{0.1cm}
\caption{\footnotesize 
Overview of the key properties designed in \benchmark~tasks. Task Decomposition (1st row): whether sub-tasks can be completed in parallel or sequentially; Observation space (2nd row): whether all agents receive the same information of task status; Workspace Overlap (3rd row): proximity between robots during execution.}
\vspace{-5mm}
\label{task-design-table} 
\end{wraptable}

%% file: text/result.tex
\vspace{-4mm}
\section{Experiments}
\vspace{-2mm}
\textbf{Overview.} We design a series of experiments using \benchmark~to validate our approach. In Section \ref{result:main}, we evaluate the task performance of \ours~compared to an oracle LLM-planner that does not use dialog, and ablate on different components of the dialog prompting in \ours. Section \ref{result:waypoint} shows empirically the benefits of LLM-proposed 3D waypoints in multi-arm motion planning. Section \ref{result:qual} contains qualitative results that demonstrate the flexibility and adaptability of \ours. Additional experiment results, such as failure analysis, are provided in Appendix \ref{appendix:more-result}.
\vspace{-3mm}
\subsection{Main Results on \benchmark}
\label{result:main}
\vspace{-3mm}
\textbf{Experiment Setup.} We use GPT-4 \cite{OpenAI2023GPT4TR} for all our main results. In addition to our method `Dialog', we set up an oracle LLM-planner `Central Plan', which is given full environment observation, information on all robots' capabilities and the same plan feedback, and prompts an LLM to plan actions for all robots at once. We also evaluate two ablations on the prompt components of \ours: first removes dialog and action history from past rounds, i.e. `Dialog w/o History'. Second removes environment feedback, i.e. `Dialog w/o Feedback', where a failed action plan is discarded and agents are prompted to continue discussion without detailed failure reasons. To offset the lack of re-plan rounds, each episode is given twice the budget of episode length.

\textbf{Evaluation Metric.}
Provided with a finite number of rounds per episode and a maximum number of re-plan attempts per round, we evaluate the task performance on three metrics: 1) task success rate within the finite rounds; 2) number of environment steps the agents took to succeed an episode, which measures the efficiency of the robots' strategy; 3) average number of re-plan attempts at each round before an environment action is executed -- this reflects the agents' ability to understand and use environment feedback to improve their plans.
Overall, a method is considered better if the task success rate is higher, it takes fewer environment steps, and requires fewer number of re-plans.

\textbf{Results.}
The evaluation results are reported in Table \ref{table:eval-result}. We remark that, despite receiving less information, dialog agents sometimes achieve comparable performance to the oracle planner. Particularly in Sort Cubes task, agents are able to find a strategy to help each other through dialog, but the oracle makes mistakes in trying to satisfy all agents' constraints at once. While removing history information or plan feedback rounds does not negatively impact performance on some tasks, full prompt that includes both achieves the best overall results. Lastly, on Pack Gocery task, the oracle planner shows better capability in waypoint planning, displaying better capability at incorporating feedback and improve on individual coordinate steps.  

\input{text/table-main-result}
\vspace{-4mm}
\subsection{Effect of LLM-proposed 3D Waypoints}
\begin{figure}[h!]
    \centering
    \includegraphics[width=0.97\textwidth]{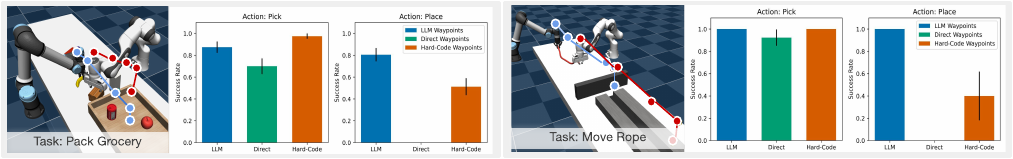}
    \vspace{-1mm}
    \caption{\footnotesize We demonstrate the utility of LLM-proposed waypoints by comparing with two alternatives: a linear waypoint path that interpolates between start and goal; `Hard-code', a predefined waypoint path that always performs top-down pick or place. 
    }
    \vspace{-6mm}
    \label{fig:waypoints-results}
\end{figure}

\vspace{-4mm}
\label{result:waypoint}
We demonstrate the utility of LLM-proposed task space waypoints. We use two tasks that were designed to have high workspace overlap, i.e. Pack Grocery and Move Rope, which require both picking and placing to complete the task. For comparison, we define a hard-coded waypoint path that performs top-down pick or place, i.e. always hovers over a gripper atop a certain height before picking an object, and moves an object up to a height above the table before moving and placing. We single-out one-step pick or place snapshots, and run multi-arm motion planning using the compared waypoints, under a maximum of 300 second planning time budget.\\
As shown in Fig.~\ref{fig:waypoints-results}, LLM-proposed waypoints show no clear benefits for picking sub-tasks, but significantly accelerate planning for placing, where collisions are more likely to happen between the arms and the desktop objects.

\subsection{Zero-shot Adaptation to Task Variations} \label{result:qual}
\vspace{-2mm}
Leveraging the zero- and few-shot ability of LLMs, \ours~demonstrates strong adaptation ability to varying task semantics, which traditionally would require modification or re-programming of a system, e.g. fine-tuning a learning-based policy. We showcase 3 main variation categories, all using Make Sandwich task in \benchmark.\\
\textbf{1. Object Initialization}: The locations of the food items are randomized, and we show the dialog agents' reasoning is robust to this variation. \\
\textbf{2. Task Goal}: The agents must stack food items in the correct order given in the sandwich recipe, and are able to coordinate sub-task strategies accordingly. \\
\textbf{3. Robot Capability}: The agents are able to exchange information on items that are within their respective reach and coordinate their plans accordingly. 

\begin{figure}[h!]
     \centering \includegraphics[width=0.9\linewidth]{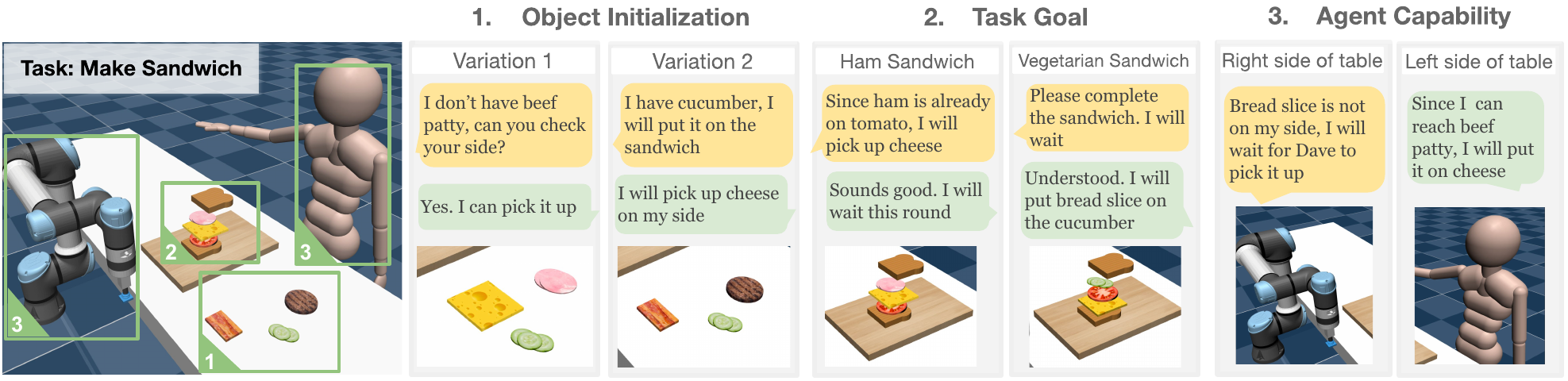}
     \caption{\footnotesize
     \ours~demonstrates strong adaptation ability to variations in task semantics. We use Make Sandwich task in \benchmark~to showcase three variation categories: 1) object initialization, i.e. randomized food items' locations on the table; 2) task goals, i.e. robots must change behavior according to different sandwich recipes; 3) agent capability, e.g. agents can only pick food items that are within their reach range. 
     }
     \vspace{-3mm}
     \label{fig:variations}
 \end{figure}

\input{text/table-real-world}
\subsection{Real-world Experiments: Human-Robot Collaboration}
\begin{figure}[h!]
     \centering \includegraphics[width=0.9\linewidth]{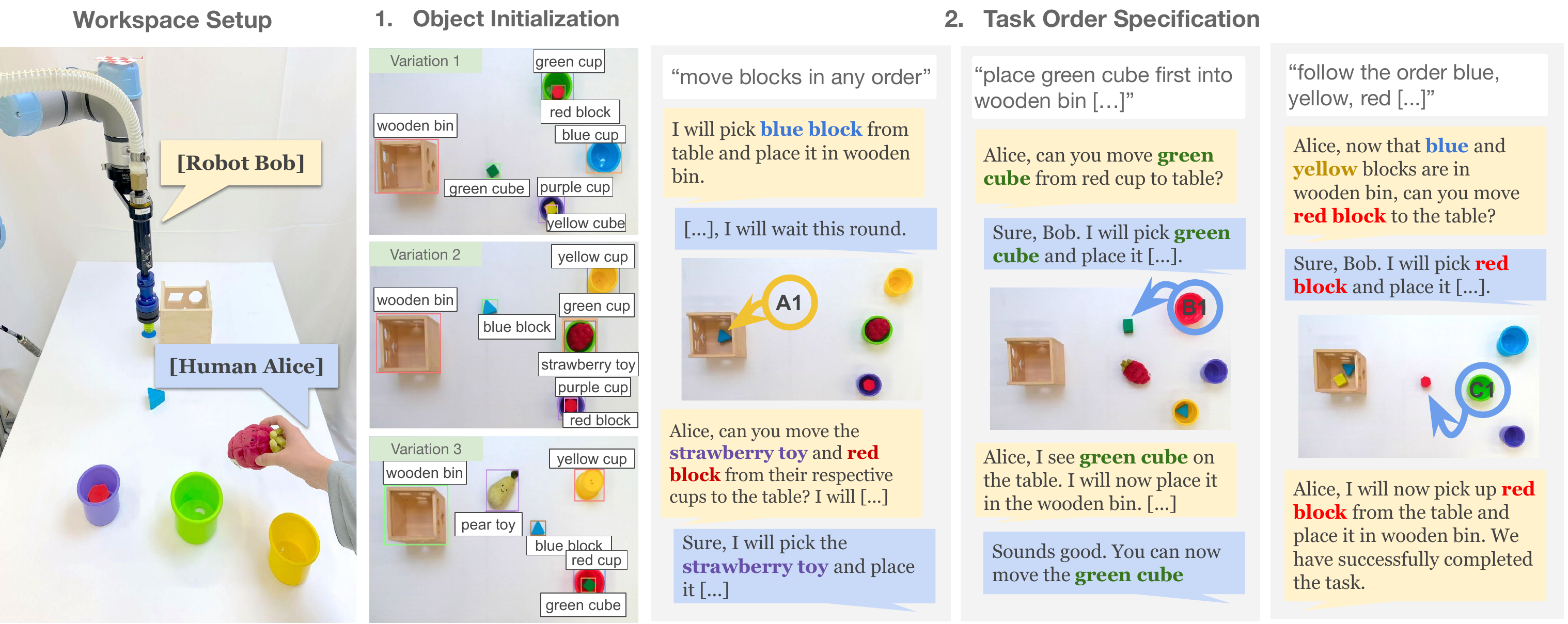}
     \caption{\footnotesize
     Real world experiments: collaborative block sorting between a robot and a human, with varying task semantics. We test two variation categories: 1) object initialization, i.e. the object locations are randomized for each episode 2) task order specification, i.e. agents must follow the specified order to move blocks.
     }
     \vspace{-6mm}
     \label{fig:real-exp}
 \end{figure}

\vspace{-2mm}
We validate \ours~in a real world setup, where a robot arm collaborates with a human to complete a sorting blocks task (Fig. \ref{fig:real-exp}). We run \ours~with the modification that only the robot agent is controlled by GPT-4, and it discusses with a human user that interacts with part of the task workspace. For perception, we use a pre-trained object detection model, OWL-ViT \cite{minderer2022simple}, to generate scene description from top-down RGB-D camera images. The task constrains the human to only move blocks from cups to the table, then the robot only picks blocks from table into wooden bin. We evaluate 2 main variation categories: 1) object initialization, i.e. initial block locations are randomized for each run (Fig.~\ref{fig:real-exp}.1); 2) task order specification, where the agents are asked to follow a fixed order to move the blocks (Fig.~\ref{fig:real-exp}.2). We also evaluate two types of human behaviors: first is an oracle human that corrects mistakes in the OWL-ViT-guided scene descriptions and the robot's responses; second is an imperfect human that provides no feedback to those errors. 

We evaluate 10 runs for each setup, see Table \ref{result:real-world} for results. We report task success rate within the finite rounds, and number of steps the agents took to succeed an episode. We remark that task performance is primarily bottle-necked by incorrect object detection from OWL-ViT, which leads to either an incorrect object being picked up and resulting in failure or no object being picked up and resulting in higher steps. See Appendix \ref{appendix:robot-prompt} for further details on the real world experiments.


%% file: text/table-main-result.tex
\begin{table}[]
\footnotesize
\begin{tabular}{@{}>{\centering\arraybackslash}p{1.3cm}>{\centering\arraybackslash}p{1.3cm}cccccc@{}}
\toprule[1pt] 
      &         & \begin{tabular}[c]{@{}c@{}}Pack\\ Grocery\end{tabular} & \begin{tabular}[c]{@{}c@{}}Arrange\\ Cabinet\end{tabular} & \begin{tabular}[c]{@{}c@{}}Sweep\\ Floor\end{tabular} & \begin{tabular}[c]{@{}c@{}}Make \\ Sandwich\end{tabular} & \begin{tabular}[c]{@{}c@{}}Sort\\ Cubes\end{tabular} & \begin{tabular}[c]{@{}c@{}}Move \\ Rope\end{tabular} \\ [7pt]\midrule[0.9pt]

\multirow{2}{*}{\begin{tabular}[c]{@{}c@{}}Central Plan\\ (oracle)\end{tabular}} & \scriptsize{Success} 
& 0.82 ± 0.06  & 0.90 ± 0.07 & 1.00 ± 0.00 & 0.96 ± 0.04 & 0.70 ± 0.10 & 0.50 ± 0.11 \\ 
& \scriptsize{step, replan}  
& 11.1, 3.9  & 4.0, 2.7 & 8.4, 2.0 & 8.8, 1.2 & 8.6, 2.6 & 2.3, 3.9  \\ \midrule[0.9pt]

\multirow{2}{*}{\begin{tabular}[c]{@{}c@{}}Dialog w/o\\ History\end{tabular}} & \scriptsize{Success} 
& 0.48 ± 0.11  & 1.00 ± 0.00   & 0.00 ± 0.00  & 0.33 ± 0.12  & 0.73 ± 0.11 & 0.65 ± 0.11 \\
& \scriptsize{step, replan} & 9.2, 3.1  & 4.2, 1.4 & N/A, 1.0 & 9.6, 1.8 & 5.8, 1.4 & 3.7, 3.1   \\ \midrule
 

\multirow{2}{*}{\begin{tabular}[c]{@{}c@{}}Dialog w/o\\ Feedback\end{tabular}}   & \scriptsize{Success} 
&  0.35 ± 0.10 & 0.70 ± 0.10 & 0.95 ± 0.05 & 0.35 ± 0.11 & 0.53 ± 0.13 & 0.45 ± 0.11 \\
 & \scriptsize{step, replan} 
 & 18.0, 1.0 & 5.9, 1.0 & 7.6, 1.0 & 12.6, 1.0 & 4.9, 1.0 & 3.4, 1.0 \\ \midrule

\multirow{2}{*}{\begin{tabular}[c]{@{}c@{}} Dialog\\ (ours)\end{tabular}}  & \scriptsize{Success} 
& 0.44 ± 0.06 & 0.75 ± 0.10 & 0.95 ± 0.05 & 0.80 ± 0.08 & 0.93 ± 0.06 & 0.65 ± 0.11   \\
 & \scriptsize{step, replan} 
 & 9.9, 3.5  & 4.7, 2.0 & 7.1, 1.0 & 10.2, 1.7 & 4.9, 1.3 & 2.5, 3.1   \\ 
 \bottomrule[1pt]
\end{tabular}
\vspace{1mm}
\caption{\footnotesize Evaluation results on \benchmark. We report averaged success rates $\uparrow$ over 20 runs per task, the average number of steps in \textit{successful} runs $\downarrow$ , and average number of re-plan attempts used across all runs $\downarrow$ .}
\label{table:eval-result}
\vspace{-5mm}
\end{table} 
 

%% file: text/table-real-world.tex

\begin{wraptable}{r}{0.4\linewidth}
\centering
\vspace{-5mm}
\setlength{\tabcolsep}{0.02cm}
\centering
\footnotesize 
\begin{tabular}{cccc}
\toprule
& & Object Init.         & Task Order           \\ \midrule
\multirow{2}{*}{\begin{tabular}[c]{@{}c@{}}Human \\ Correction\end{tabular}}   
& Success           &    9/10    &    8/10   \\ 
& Step      &    5.3  &                5.5      \\ \midrule
\multirow{2}{*}{\begin{tabular}[c]{@{}c@{}}Imperfect \\ Human\end{tabular}} 
& Success           &    7/10    &    6/10   \\ 
& Step      &    5.6  &                5.2      \\ \bottomrule
\end{tabular}
\caption{\footnotesize Real world experiment results. We report number of successes and average number of steps in sucessful runs.}
\vspace{-3mm}
\label{result:real-world} 
\end{wraptable}

%% file: text/reasoning-dataset.tex
\vspace{-2.5mm}
\section{Multi-Agent Representation and Reasoning Dataset} 
\vspace{-3mm}
In addition to our main experimental results, we curate a text-based dataset, \benchmark-Text, to evaluate an LLM's agent representation and task reasoning ability. This dataset aligns LLM with desirable capabilities in multi-agent collaboration, without requiring robotic environment interaction. It builds on data from our evaluation on \benchmark, and contains a series of additional questions that are more open-ended and go beyond simply finding the next best action plan.

\subsection{Dataset Overview}
\vspace{-3mm}
This dataset contains yes/no, multiple-choice or short question-answering questions, spanning a range of different reasoning abilities (see Appendix \ref{appendix:reasoning} for more details): \\
\textbf{Self-knowledge} evaluates how well the agent establishes its identity under a given task context, divided into two categories: 1) understanding an agent's own capability (e.g. which objects/area are not reachable); 2) memory retrieval, i.e. inferring information from past dialog and actions.\\
\textbf{Communication Skills} evaluates an agent's ability to effectively exchange information and drive a discussion into an agreeable plan. The questions ask an LLM to 1) choose appropriate response to other agents' questions; 2) choose appropriate inquiries to other agents.\\
\textbf{Adaptation} evaluates adaptation to unexpected situations that were not specified in context. We use a subset of \benchmark~ tasks to design unexpected occurrences, either regarding task state (e.g. a missing object) or a response from another agent, and ask an LLM agent to choose the best response. See below for an example question: two agents make a sandwich together; one agent is informed of a broken gripper and must infer that the sandwich can actually be completed without any item from its side of the table.

\promptblock{\footnotesize   
You are a robot Chad collaborating with Dave ...\blank{[task context]}
Your gripper is not working. What should you say to Dave? 
Select exactly one option from below. \newline 
A: Sorry Dave, we can't complete the task anymore, my gripper is broke.\newline
B: Let's stop. The recipe needs ham but Dave can't reach my side and my gripper is not functioning.\newline
\textcolor{dark-green}{C: Dave, go ahead and finish the sandwich without me, there isn't anything we need on my side anyway.} 
}
\vspace{-3mm}
\subsection{Evaluation Results}
\vspace{-3mm}
\textbf{Setup.} All questions are designed to have only one correct answer, hence we measure the average accuracy in each category. We evaluate GPT-4 (OpenAI), GPT-3.5-turbo (OpenAI), and Claude-v1 (Anthropic\cite{claude}). For GPT-4, we use two models marked with different time-stamps, i.e. 03/14/2023 and 06/13/2023. Results are summarized in Table \ref{table:reasoning}: we observe that, with small performance variations between the two versions, GPT-4 leads the performance across all categories. We remark that there is still a considerable gap from fully accurate, and hope this dataset will be useful for improving and evaluating language models in future work. \\
\textbf{Qualitative Results.} We observe GPT-4 is better at following the instruction to formulate output, whereas GPT-3.5-turbo is more prone to confident and elongated wrong answers. See below for an example response from an agent capability question (the prompt is redacted for readability).\\
\promptblock{\footnotesize You are robot Chad .. \blank{[cube-on-panel locations...]}. You can reach: \blank{[panels]} 
\newline Which cube(s) can you reach? \blank{[...]} Answer with a list of cube names, answer None if you can't reach any. \newline
\textcolor{blue}{Solution: None} \newline
\llmcompletion{GPT-4: None} 
\yellowcompletion{Claude-v1: yellow\_trapezoid} \newline
\greencompletion{GPT-3.5-turbo: At the current round, I can reach the yellow\_trapezoid cube on panel3.}
}
\input{text/table-reasoning-results}

%% file: text/table-reasoning-results.tex
            

\begin{table}[]
\centering
\vspace{-3mm}
\footnotesize
\begin{tabular}{@{}lccccc@{}}
\toprule[1pt]
 & \multicolumn{2}{c}{Self-knowledge} & \multicolumn{2}{c}{Communication} & Adaptation \\
 & Capability         & Memory        & Inquiry         & Respond         &            \\ \midrule
            
GPT-4-0314    & 0.67 ± 0.06   & 0.84 ± 0.06 & \textbf{0.79 ± 0.06} & 0.83 ± 0.04  & 0.68 ± 0.08 \\  \midrule
GPT-4-0613    & \textbf{0.68 ± 0.06}   & \textbf{0.91 ± 0.04} & 0.57 ± 0.08 & \textbf{0.86 ± 0.03} & \textbf{0.71 ± 0.08} \\  \midrule
GPT-3.5-turbo & \textbf{0.68 ± 0.06}   & 0.59 ± 0.07 & 0.48 ± 0.08 & 0.30 ± 0.05  & 0.58 ± 0.09 \\  \midrule
Claude-v1     & 0.37 ± 0.06   & 0.70 ± 0.07 & 0.55 ± 0.08 & 0.60 ± 0.05  & 0.65 ± 0.09 \\   \bottomrule
\end{tabular}
\vspace{2mm}
\caption{\footnotesize Evaluation results on the multi-agent LLM reasoning dataset. We measure the question-answering accuracy on each test category and compare performance of four different models.
}
\vspace{-8mm}
\label{table:reasoning}
\end{table}  

%% file: text/relatedwork.tex
\vspace{-2mm}
\section{Related Work}
\vspace{-3mm}
\textbf{LLMs for Robotics.}
An initial line of prior work uses LLMs to select skill primitives and complete robotic tasks, such as SayCan \cite{ahn2022i}, Inner Monologue \cite{Huang2022InnerME}, which, similarly to ours, uses environment feedback to in-context improve planning. Later work leverages the code-generation abilities of LLMs to generate robot policies in code format, such as CaP \cite{codeaspolicies2022}, ProgGPT \cite{singh2022progprompt} and Demo2Code \cite{wang2023demo2code}; or longer programs for robot execution such as TidyBot \cite{wu2023tidybot} and Instruct2Act \cite{huang2023instruct2act}. 
Related to our use of motion-planner, prior work such as Text2Motion \cite{lin2023text2motion}, AutoTAMP \cite{chen2023autotamp} and LLM-GROP \cite{ding2023llmgrop, zhang2022grop} studies combining LLMs with traditional Task and Motion Planning (TAMP).
Other work explores using LLMs to facilitate human-robot collaboration \cite{cui2023no}, to design rewards for reinforcement learning (RL) \cite{kwon2023reward}, and for real-time motion-planning control in robotic tasks \cite{yu2023synthesis}. 
While prior work uses single-robot setups and single-thread LLM planning, we consider multi-robot settings that can achieve more complex tasks, and use dialog prompting for task reasoning and coordination. 

\textbf{Multi-modal Pre-training for Robotics.}
LLMs' lack of perception ability bottlenecks its combination with robotic applications. One solution is to pre-train new models with both vision, language and large-scale robot data: the multi-modal pre-trained PALM-E \cite{driess2023palm} achieves both perception and task planning with a single model; Interactive Language \cite{lynch2022interactive} and DIAL \cite{xiao2022robotic} builds a large dataset of language-annotated robot trajectories for training generalizable imitation policies. Another solution is to introduce other pre-trained models, mainly vision-language models (VLMs) such as CLIP \cite{radford2021clip}). In works such as Socratic Models \cite{zeng2022socratic}, Matcha \cite{zhao2023matcha}, and \citet{kwon2023grounded}, LLMs are used to repeatedly query and synthesize information from other models to improve reasoning about the environment. While most use zero-shot LLMs and VLMs, works such as CogLoop \cite{jin2023alphablock} also explores fine-tuning adaptation layers to better bridge different frozen models. Our work takes advantage of simulation to extract perceptual information, and our real world experiments follow prior work \cite{liang2022code,singh2022progprompt,wu2023tidybot} in using pre-trained object detection models \cite{minderer2022simple} to generate scene description.

\textbf{Dialogue, Debate, and Role-play LLMs.}
Outside of robotics, LLMs have been shown to possess the capability of representing agentic intentions \cite{andreas2022language} and behaviors, which enables multi-agent interactions in simulated environments such as text-based games \cite{schlangen2023dialogue, chalamalasetti2023clembench} and social sandbox \cite{park2023generative, Li2023CAMELCA, liu2023training}. Recent work also shows a dialog or debate style prompting can improve LLMs' performance on human alignment \cite{irving2018ai} and a broad range of goal-oriented tasks \cite{nair2023dera,Liang2023EncouragingDT,du2023improving}. While prior work focuses more on understanding LLM behaviors or improve solution to a single question, our setup requires planning separate actions for each agent, thus adding to the complexity of discussion and the difficulty in achieving consensus.


\textbf{Multi-Robot Collaboration and Motion Planning.} Research on multi-robot manipulation has a long line of history \cite{Koga1994OnMM}. A first cluster of work studies the low-level problem of finding collision-free motion trajectories. Sampling-based methods are a popular approach \cite{karaman2011samplingbased}, where various algorithmic improvements have been proposed \cite{dobson2015planning}. Recent work also explored learning-based methods \cite{ha2020multiarm} as alternative. While our tasks are mainly set in static scenes, much work has also studied more challenging scenarios that require more complex low-level control, such as involving dynamic objects \cite{salehian2016coordinated} or closed-chain kinematics \cite{link-closed-chains,xian2017closed}. A second cluster of work focuses more on high-level planning to allocate and coordinate sub-tasks, which our work is more relevant to. While most prior work tailor their systems to a small set of tasks, such as furniture assembly \cite{furniture}, we highlight the generality of our approach to the variety of tasks it enables in few-shot fashion.

%% file: text/appendix.tex
\newpage
\section{\benchmark}\label{appendix:benchmark}
\subsection{Overview}
\benchmark~is built with MuJoCo \cite{mujoco} physics engine. The authors would like to thank the various related open-source efforts that greatly assisted the development of \benchmark~tasks: DMControl \cite{dm_control}, 
 Menagerie\cite{menagerie2022github}, and MuJoCo object assets from \citet{dasari2023pgdm}. The sections below provide a detailed documentation for each of the 6 simulated collaboration tasks.
 
\subsection{Task: Sweep Floor}
\begin{wrapfigure}{Rt}{0.3\textwidth} 
\vspace{-13mm}
\centering
\includegraphics[width=0.3\textwidth]{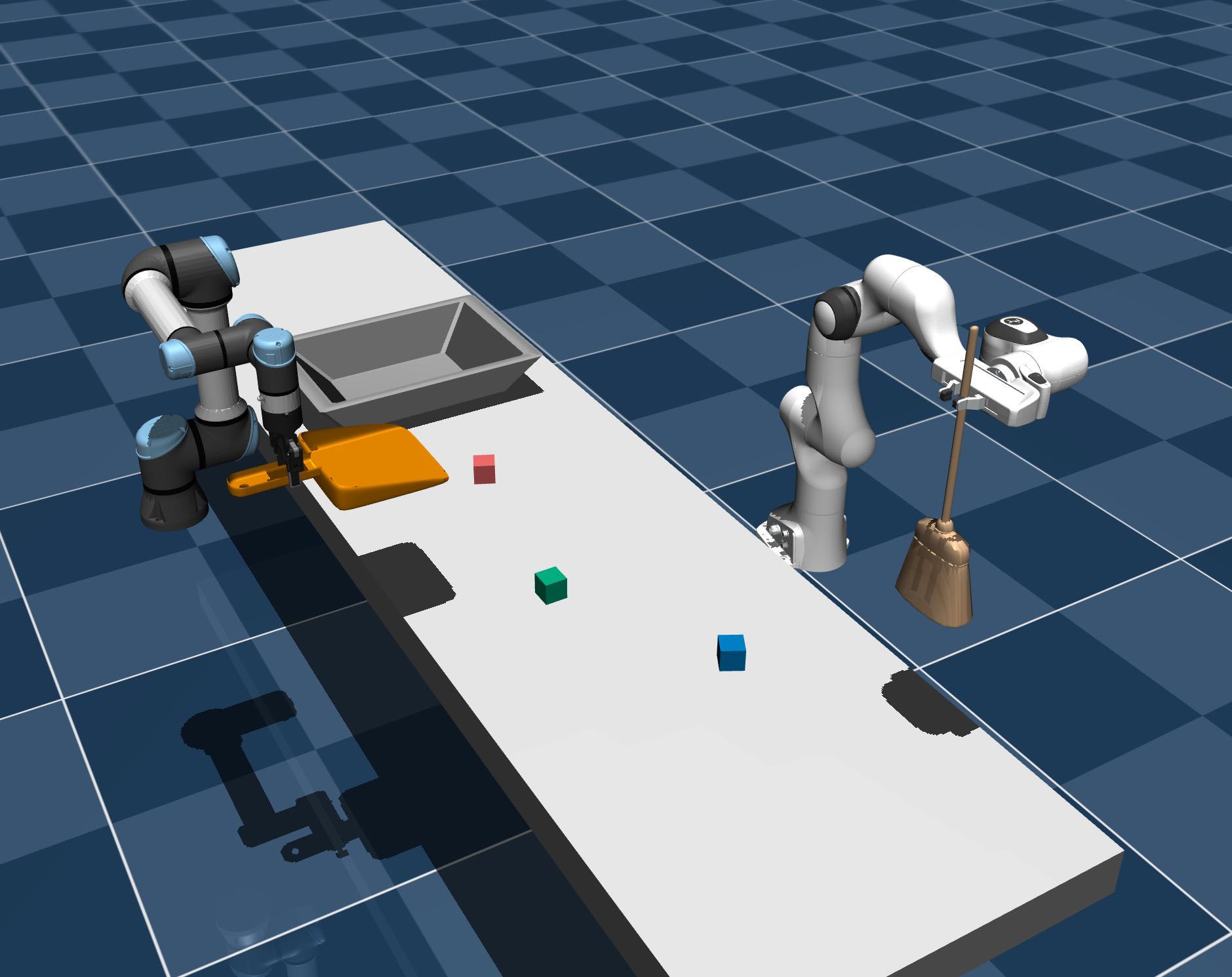}
\caption{Sweep Floor task }
\label{fig:sweep} 
\vspace{-5mm}
\end{wrapfigure}
\textbf{Task Description.} 2 Robots bring a dustpan and a broom to opposite sides of each cube to sweep it up, then the robot holding dustpan dumps cubes into a trash bin. 

\textbf{Agent Capability.} Two robots stand on opposite sides of the table:
\vspace{-6mm}
\begin{enumerate}[leftmargin=*]
    \item UR5E with robotiq gripper (`Alice'): holds a dustpan
    \item Franka Panda (`Bob'): holds a broom
\end{enumerate}
\vspace{-2mm}
\textbf{Observation Space.}
1) cube locations: a. on table; b. inside dustpan; c. inside trash bin; 2) robot status: 3D gripper locations

\textbf{Available Robot Skills.}
1) MOVE [target]: target can only be a cube; 2) SWEEP [target]: moves the groom so it pushes the target into dustpan; 3) WAIT; 4) DUMP: dump dustpan over the top of trash bin.  

\subsection{Task: Make Sandwich}
\begin{wrapfigure}{Rt}{0.3\textwidth} 
\vspace{-16mm}
\centering
\includegraphics[width=0.3\textwidth]{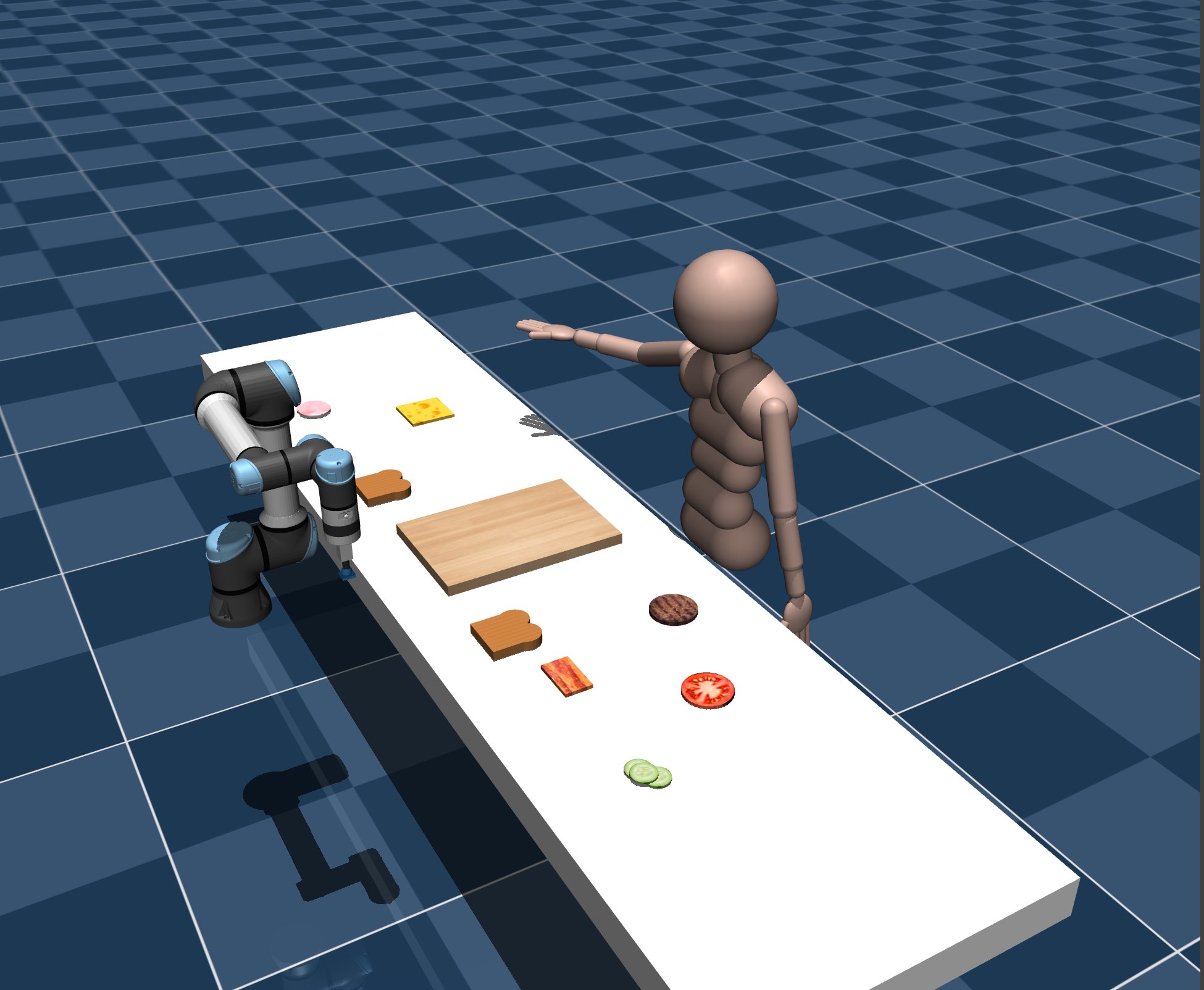}
\caption{Make Sandwich task }
\label{fig:sandwich} 
\end{wrapfigure}
\textbf{Task Description.} 2 Robots make a sandwich together, each having access to a different set of ingredients. They must select the required items and take turns to stack them in the correct order. 

\textbf{Agent Capability.} Two robots stand on opposite sides of the table:
\vspace{-6mm}
\begin{enumerate}[leftmargin=*]
    \item UR5E with suction gripper (`Chad'): can only reach right side 
    \item Humanoid robot with suction gripper (`Dave'): can only reach left side  
\end{enumerate}
\vspace{-2mm}
\textbf{Observation Space}
1) the robot's own gripper state (either empty or holding an object); 2) food items on the robot's own side of the table and on the cutting board.

\textbf{Available Robot Skills.} 1) PICK [object]; 2) PUT [object] on [target]; WAIT

\subsection{Task: Sort Cubes}
\begin{wrapfigure}{Rt}{0.3\textwidth} 
\vspace{-15mm}
\centering
\includegraphics[width=0.3\textwidth]{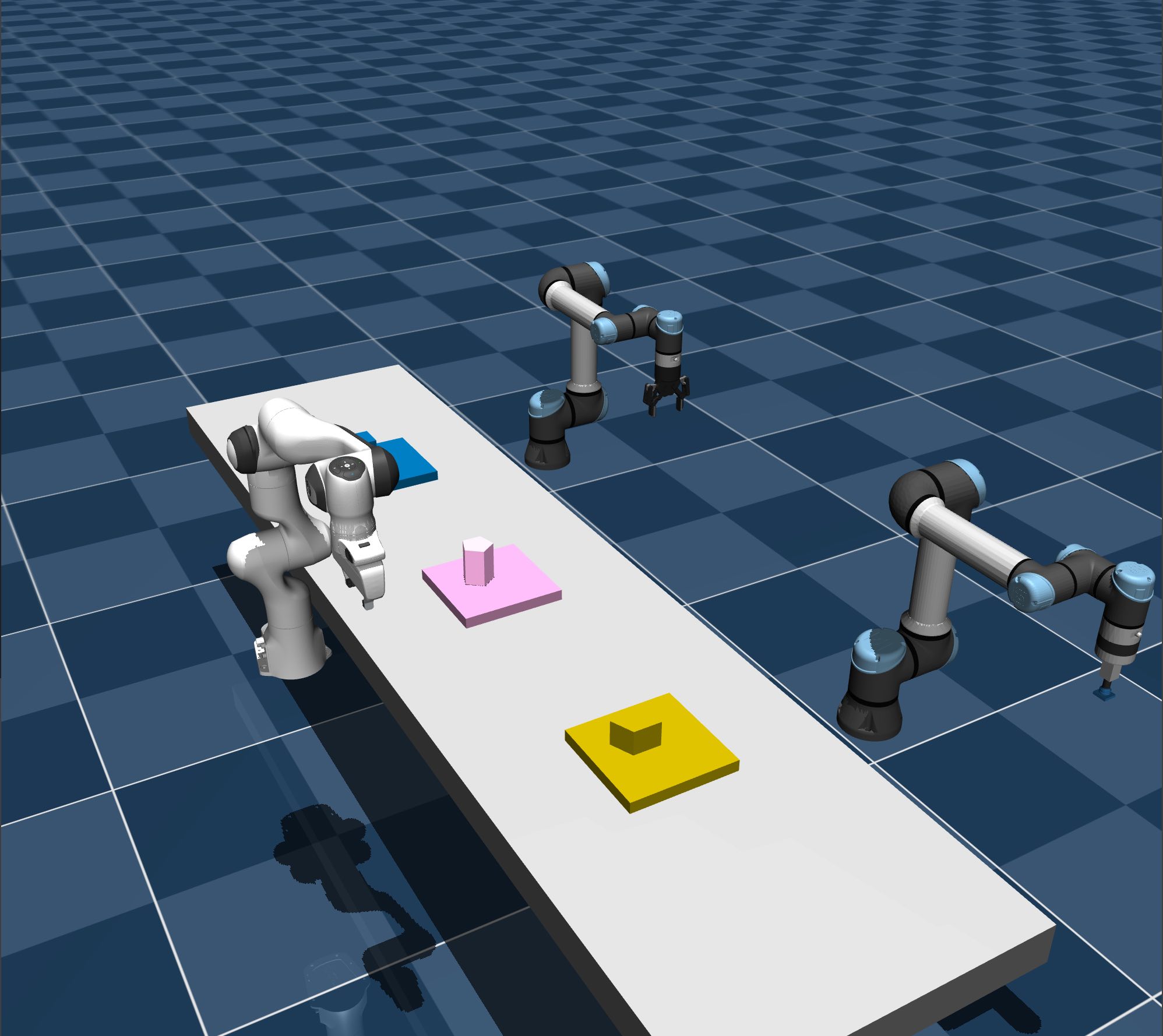}
\caption{Sort Cubes task }
\label{fig:sort} 
\vspace{-11mm}
\end{wrapfigure}
\textbf{Task Description.} 3 Robots sort 3 cubes onto their corresponding panels. The robots must stay within their respective reach range, and help each other to move a cube closer. 

\textbf{Agent Capability.} Three robots each responsible for one area on the table
\vspace{-3mm}
\begin{enumerate}[leftmargin=*]
    \item UR5E with robotiq gripper (`Alice'): must put blue square on panel2, can only reach: panel1, panel2, panel3.
    \item Franka Panda (`Bob'): must put pink polygon on panel4, can only reach: panel3, panel4, panel5.
    \item UR5E with suction gripper (`Chad'): must put yellow trapezoid on panel6, can only reach: panel5, panel6, panel7.
\end{enumerate}
\vspace{-2mm}
\textbf{Observation Space}
1) the robot's own goal,
2) locations of each cube.

\textbf{Available Robot Skills.} 1) PICK [object] PLACE [panelX]; 2) WAIT

\subsection{Task: Pack Grocery}
\begin{wrapfigure}{Rt}{0.3\textwidth} 
\vspace{-18mm}
\centering
\includegraphics[width=0.3\textwidth]{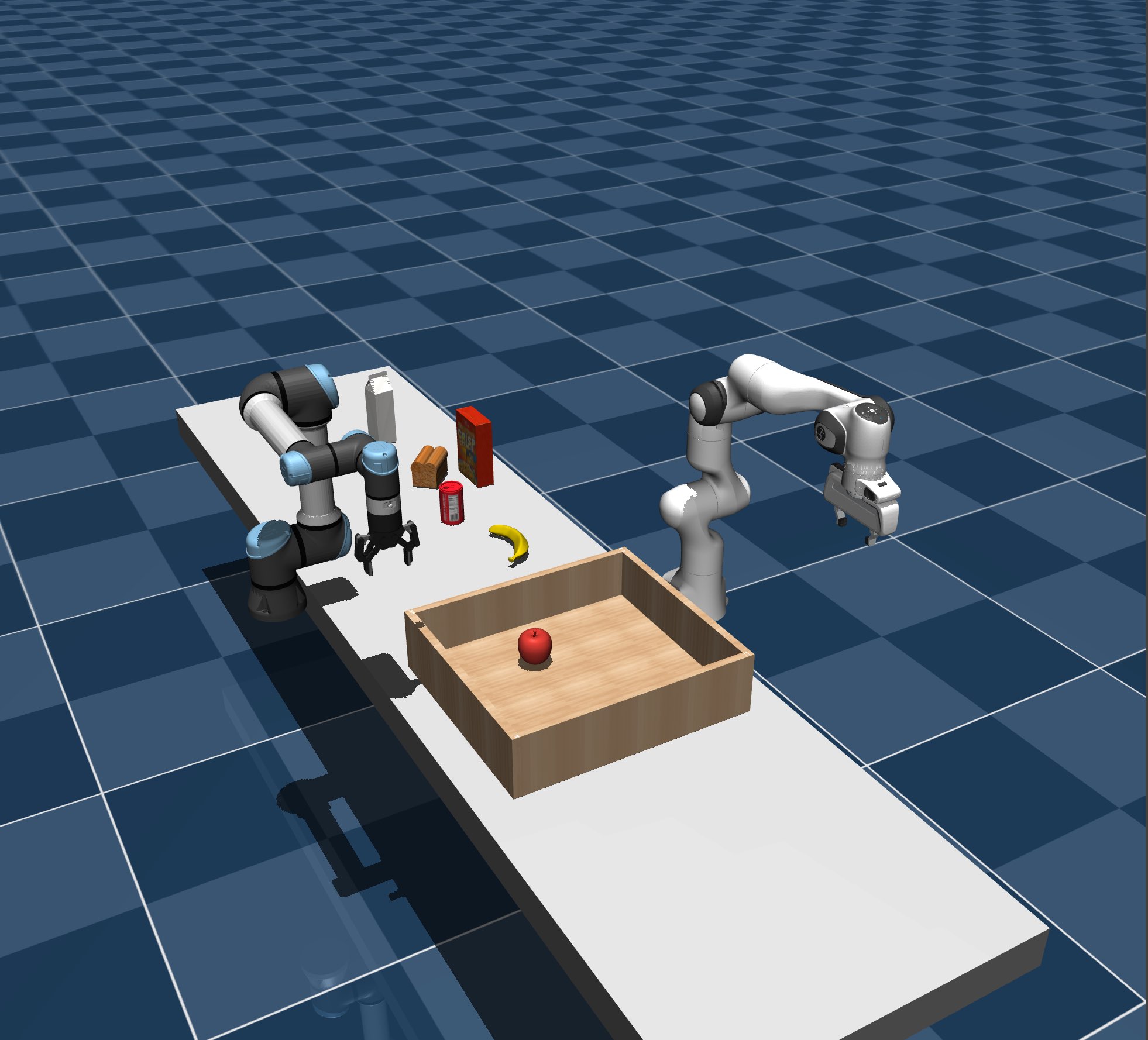}
\caption{Pack Grocery task }
\label{fig:pack}
\vspace{-8mm}
\end{wrapfigure}
\textbf{Task Description.} 2 Robots pack a set of grocery items from the table into a bin. The objects are in close proximity and robots must coordinate their paths to avoid collision.

\textbf{Agent Capability.} Two robots on opposite sides of table
\vspace{-3mm}
\begin{enumerate}[leftmargin=*]
    \item UR5E with robotiq gripper (`Alice'): can pick and place any object on the table
    \item Franka Panda (`Bob'): can pick and place any object on the table 
\end{enumerate}
\vspace{-2mm}
\textbf{Observation Space} 1) robots' gripper locations, 2) locations of each object, 3) locations of all slots in the bin.

\textbf{Available Robot Skills.} (must include task-space waypoints) 1) PICK [object] PATH [path]; 2) PLACE [object] [target] PATH [path]

\subsection{Task: Move Rope}
\begin{wrapfigure}{Rt}{0.3\textwidth} 
\vspace{-15mm}
\centering
\includegraphics[width=0.3\textwidth]{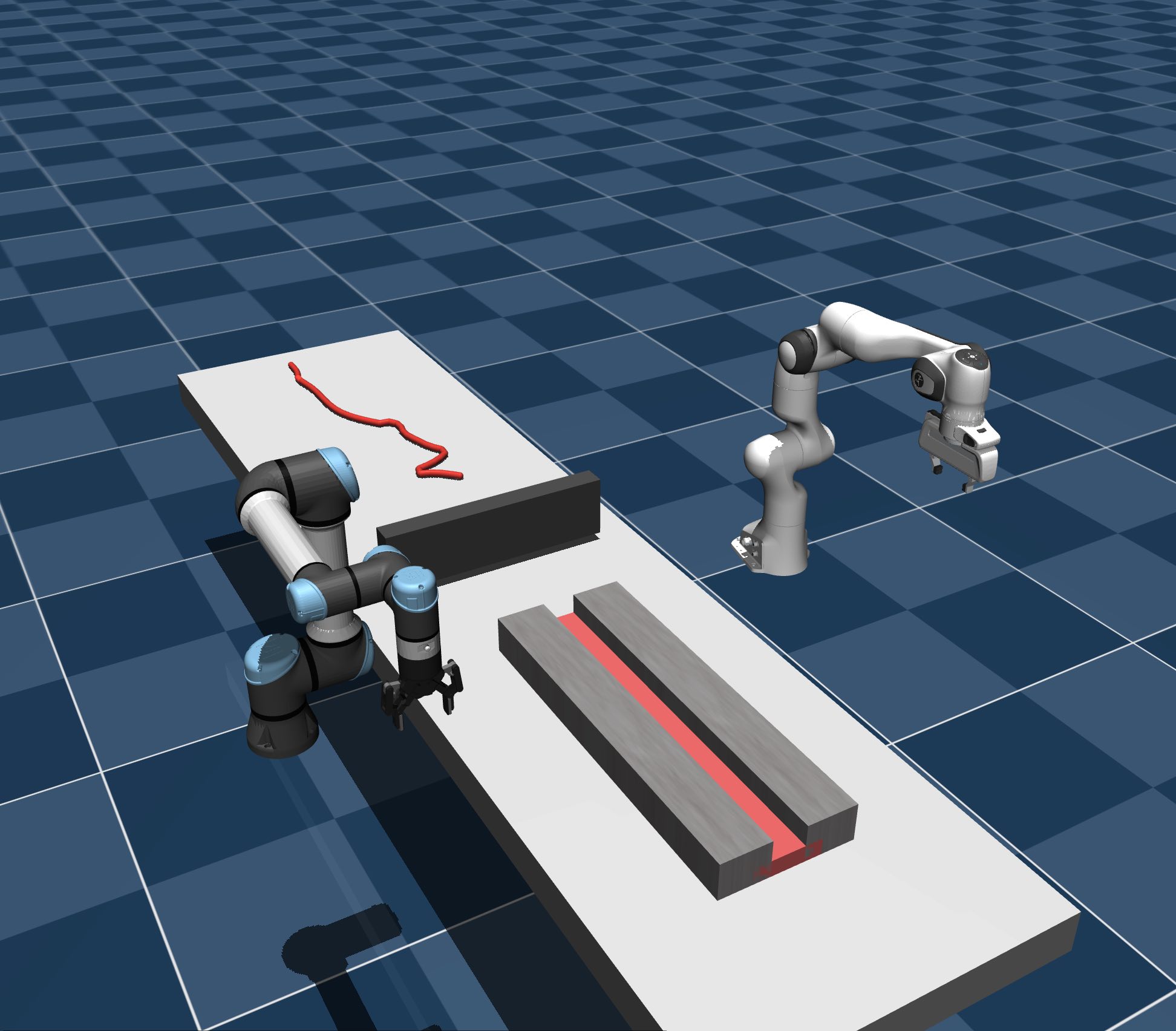}
\caption{Move Rope task }
\label{fig:rope} 
\vspace{-5mm}
\end{wrapfigure}
\textbf{Task Description.} 2 robots lift a rope together over a wall and place it into a groove. They must coordinate their grippers to avoid collision.

\textbf{Agent Capability.} Two robots on opposite sides of table
\vspace{-3mm}
\begin{enumerate}[leftmargin=*]
    \item UR5E with robotiq gripper (`Alice'): can pick and place any end of the rope within its reach
    \item Franka Panda (`Bob'): can pick and place any end of the rope within its reach
\end{enumerate}
\vspace{-2mm}
\textbf{Observation Space} 1) robots' gripper locations, 2) locations of rope's front and end back ends; 3) locations of corners of the obstacle wall; 4) locations of left and right ends of the groove slot.

\textbf{Available Robot Skills.} (must include task-space waypoints) 1) PICK [object] PATH [path]; 2) PLACE [object] [target] PATH [path]

\subsection{Task: Arrange Cabinet}
\begin{wrapfigure}{Rt}{0.3\textwidth} 
\vspace{-15mm}
\centering
\includegraphics[width=0.3\textwidth]{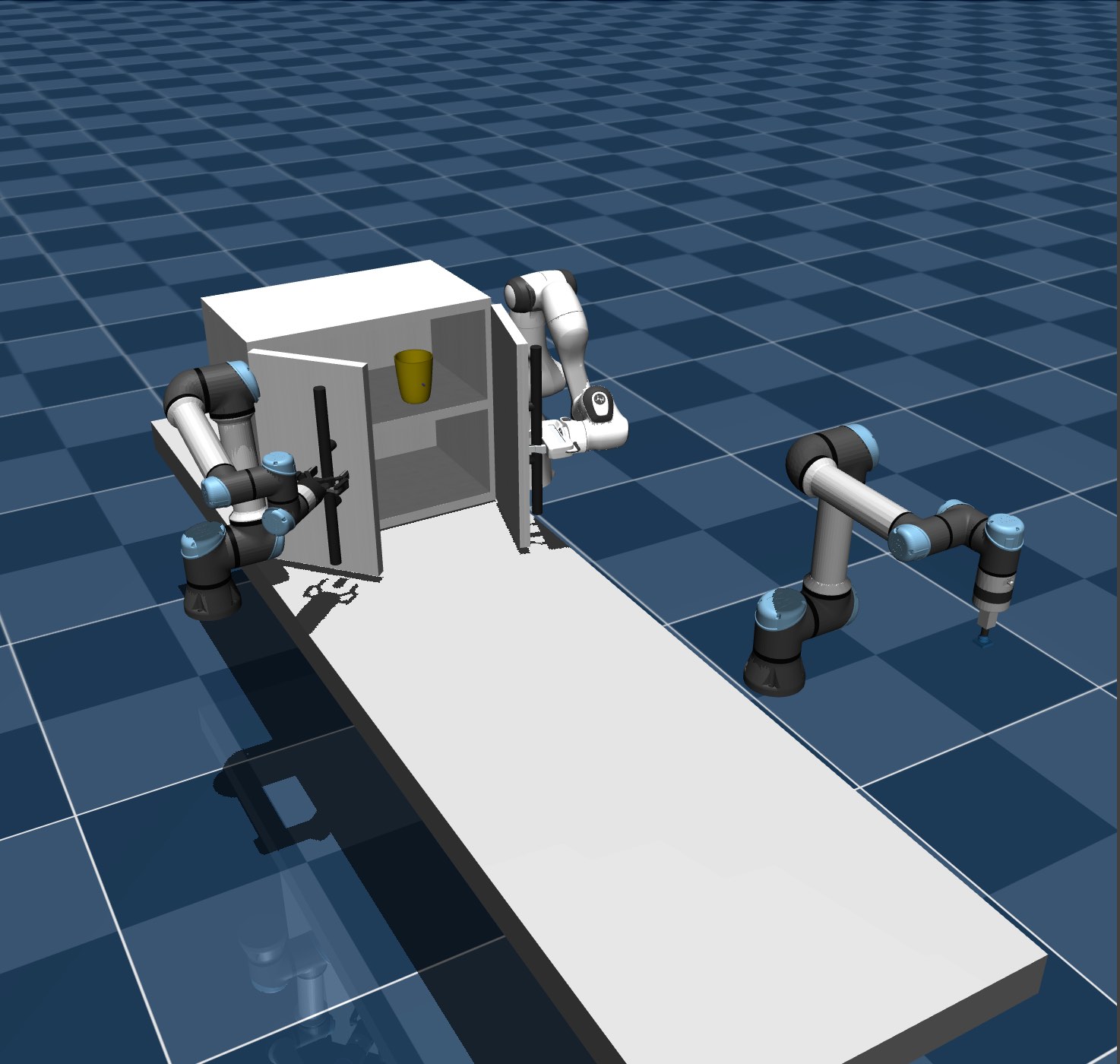}
\caption{Arrange Cabinet task }
\label{fig:sort} 
\vspace{-11mm}
\end{wrapfigure}
\textbf{Task Description.} 3 robots, two of them each hold one side of the cabinet door open, while the third robot takes the cups out and place them onto the correct coasters.

\textbf{Agent Capability.} Three robots, one on left side of the table, two on right side of table
\vspace{-3mm}
\begin{enumerate}[leftmargin=*]
    \item UR5E with robotiq gripper (`Alice'): stands on left side, can only reach left cabinet door
    \item Franka Panda (`Bob'): : stands on right side, can only reach right cabinet door
    \item UR5E with suction gripper (`Chad'):: stands on right side, can reach right cabinet door and cups and mugs inside the cabinet.
\end{enumerate}
\vspace{-2mm}
\textbf{Observation Space}
1) locations cabinet door handles; 2) each robot's reachable objects, unaware of other robot's reach range.

\textbf{Available Robot Skills.} 1) PICK [object]; 2) OPEN [one side of door handle]; 3) WAIT; 3) PLACE [object] [target] 

\newpage
\section{Details on LLM Prompting}
We use a separate query call for every agent's individual response in a dialog, see the text box below for a redacted example of an agent's prompt:

\promptblock{\footnotesize
You are robot \blank{[agent name]}, collaborating with \blank{[other agent(s)]} to \blank{[task context]}.\\
You can \textcolor{orange}{[agent capability]}.\\
Previously: \textcolor{dark-green}{[round history]}\\
At current round: \textcolor{dark-green}{[current observation]}\\
Discuss with \blank{[other agent(s)]} to coordinate and complete the task together.\textcolor{dark-green}{[communication instruction]}.\\
Never forget you are \blank{[agent name]}!
Respond very concisely by \blank{[response format]} \newline 
Previous chat:  \blank{[dialog from previous re-plan rounds]}\\
This proposed plan failed: \textcolor{dark-green}{[plan feedback]} 
Current chat:  \blank{[dialog from current round]} \\
Your response is: \\
\llmcompletion{I am Alice, ...}
}

\begin{algorithm}
\caption{Multi-agent dialog for collaboration}
\label{alg:method}
\begin{algorithmic} 
\Require agent $u^1,...,u^N$, task horizon $T$; 
\Require max number of re-plans $K$, max number of dialog per round $M$,
\Require history buffer $H$; feedback buffer $F$
\State $t \gets 0$ 
\State $o_t \gets $ env.reset()
\State $H$.empty()
\While{$t < T$}
    \State $F$.empty()
    \While{len($F$)$< K$}  
        \State dialog, plan $\gets$ PromptDialogs(H, F, $o_t$, $u^n$)
    \State plan-valid, feedback $\gets$ GiveFeedback(plan)
    \If{plan-valid} 
        \State final-plan $\gets$ parsed-plan
        \State \textbf{break} 
    \EndIf
    \State $F$.append(feedback)
\EndWhile  

\If{plan-valid} 
    \State $\sigma_t$ $\gets$ MotionPlanner($o_t$, final-plan)
    \State $o_{t+1}, r_{t+1} \gets$ env.step($\sigma_t$)
    \If{ $r_{t+1} > 0$}
        \State \textbf{break}
    \EndIf
\EndIf
\State $H$.append(dialog)
\State $t \gets t+1$
\EndWhile
\end{algorithmic}
\end{algorithm} 
\input{text/appendix-more-results}

\newpage
\input{text/appendix-toy-example}
\newpage
\input{text/appendix-reasoning-dataset}

%% file: text/appendix-more-results.tex
\section{Additional Experiment Results} \label{appendix:more-result}
\vspace{-2mm}
\subsection{Failure Analysis on \benchmark}
\vspace{-3mm}
We provide example failure scenarios observed in the agent dialog and environment interaction. 

\textbf{Factual Errors in Dialog.} We observe that when one LLM-generated agent makes a mistake, it could sometimes cause subsequent dialog to build on the wrong reasoning. See an example below: one agent (Alice) wrongfully decided the task is complete, and other agents repeat this wrong claim and choose to wait for multiple rounds while the task is, in fact, not finished.
\promptblock{\small
\textcolor{blue}{[Alice]: \commenttext{[...]} we have successfully completed our task. [...], let's wait for further instructions.} \newline
\textcolor{dark-yellow}{[Bob]: I agree that we have successfully completed our task. Great teamwork, everyone! Let's wait \commenttext{[...]}}\newline
\textcolor{red}{[Chad]: I concur that we have accomplished our goal. Excellent teamwork! Let's wait \commenttext{[...]}.} 
}

\textbf{Errors in Motion Trajectory Execution}
Due to the open-loop nature of the execution, small errors in a motion trajectory could lead to unexpected errors, e.g. knocking of an object by accident.  

\vspace{-1mm}
\subsection{Real World Experiment Setup}
\label{appendix:robot-prompt}
\vspace{-3mm}
The robot agent is a 6DoF UR5E arm with suction gripper, and dialog is enabled by querying a GPT-4 model to respond as agent `Bob', who is discussing with a human collaborator `Alice'. The human user provides text input to engage in the dialog, and arranges cubes on the same tabletop. For perception, we use top-down RGB-D image from an Azure Kinect sensor.

See the text below for an example of the robot's prompt:

\promptblock{\small 
\commenttext{==== System Prompt ====} \\
\texttt{[}Action Options\texttt{]}\\
1) PICK \texttt{<}obj\texttt{>} PLACE \texttt{<}target\texttt{>}: robot Bob must decide which block to PICK and where to PLACE. To complete the task, Bob must PLACE all blocks in the wooden bin.\\
2) WAIT: robot Bob can choose to do nothing, and wait for human Alice to move blocks from inside cups to the table.\\
\\
\texttt{[}Action Output Instruction\texttt{]}\\
First output `EXECUTE\\', then give exactly one ACTION for the robot.\\
Example\texttt{\#}1: `EXECUTE\\NAME Bob ACTION PICK green\texttt{\_}cube PLACE wooden\texttt{\_}bin\\'
Example\texttt{\#}2: `EXECUTE\\NAME Bob ACTION WAIT\\'\\
\\
You are a robot called Bob, and you are collaborating with human Alice to move blocks from inside cups to a wooden bin. \\
You cannot pick blocks when they are inside cups, but can pick blocks when they are on the table. Alice must help you by moving blocks from inside cups to the table.\\
You must WAIT for Alice to move blocks from inside cups to the table, then you can PICK blocks from the table and PLACE them in the wooden bin.\\
\commenttext{[mention task order specification]}\\
Talk with Alice to coordinate and decide what to do.\\
\\
At the current round:\\
\commenttext{[object descriptions from observation]}\\
Think step-by-step about the task and Alice's response.\\
Improve your plans if given \texttt{[}Environment Feedback\texttt{]}.\\ 
Propose exactly one action for yourself at the current round, select from \texttt{[}Action Options\texttt{]}.\\
End your response by either: 1) output PROCEED, if the plans require further discussion; 2) If everyone has made proposals and got approved, output the final plan, must strictly follow \texttt{[}Action Output Instruction\texttt{]}!        
}

\promptblock{\small 
\commenttext{==== User Prompt ====} \\
You are Bob, your response is:\\
\commenttext{response from GPT-4:} \\
\llmcompletion{EXECUTE}\\
\llmcompletion{NAME Bob ACTION ...}
}

%% file: text/appendix-toy-example.tex
\section{Toy Example on LLM for 3D Path Planning}
\label{appendix:toy-problem}
\subsection{Experiment Setup}
We use the chatCompletion mode in GPT-4. At each evaluation run, we randomly sample a new set of obstacles and agents' start-goal locations. Each run is given up to 5 attempts: using the same system prompt, which describes the task and output instructions, and a user prompt, which describes the current grid layout. If a first attempt fails, the feedback from previous plans are appended to the user prompt at later attempts, until reaching the max number of attempts. See below for the prompts and GPT-4's response (some coordinate tuples are omitted (marked as ``omitted") due to limited space).

\promptblock{\small 
\commenttext{==== System Prompt ====} \\
Plan paths for agents to navigate a 3D grid to reach their respective goals and avoid collision.\\
You are given: \\
1) a list of obstacle coordinates (x, y, z): locations of the obstacle grid cells, agents must avoid them.\\
2) a list of [([name], [init], [goal]) tuples], [init] and [goal] are 3D coordinates of the initial position and goal position of agent named [name]. \\
3) a previous plan, if any, and why it failed. Analyze this information and re-plan a collision-free path. \\ 
How to plan a path: \\
1) Make sure each path does not touch any obstacle, ... \commenttext{[planning instructions]} \\
\commenttext{[instruction on output format]}\\  
}

\promptblock{\small 
\commenttext{==== User Prompt ====} \\
At the current step: Grid size: 10 x 10 x 10 \\
Obstacles:(4, 9, 6) (0, 2, 0)... \commenttext{[list of all obstacle 3D locations]}\\
Agent Alice init: (7, 6, 2) goal: (5, 3, 2) \\
Agent Bob init: (9, 9, 5) goal: (7, 1, 4) \\
Agent Chad init: (1, 0, 0) goal: (9, 3, 6) \\
Agent Dave init: (9, 1, 1) goal: (7, 1, 0) \\
Feedback: this previous plan failed:\\ 
\commenttext{[past plan, omitted]} \\
Some steps in this plan are not exactly 1 step away from each other: Bob: (7, 4, 5), (7, 1, 5); \\
Use this information to try again, update this plan so it has collision-free, strictly one-step-apart paths. \\
Feedback: this previous plan failed: \\
...\commenttext{[another feedback, same format as above]}\\ 
\\
Your reasoning and plan is:\\
\commenttext{response from GPT-4:} \\
\llmcompletion{PLAN}\\
\llmcompletion{NAME Alice PATH [(7, 6, 2), (6, 6, 2), ...\commenttext{[omitted]} ] } \\
\llmcompletion{NAME Bob PATH [(9, 9, 5), (8, 9, 5), ...\commenttext{[omitted]} ] } \\
\llmcompletion{NAME Chad PATH [(1, 0, 0), (2, 0, 0), ...\commenttext{omitted} ] } \\
\llmcompletion{NAME Dave PATH [(9, 1, 1), (8, 1, 1), (7, 1, 1), (7, 1, 0)] } 
}
\subsection{Full Experiment Results}
\begin{figure}
    \centering
    \includegraphics[width=0.7\linewidth]{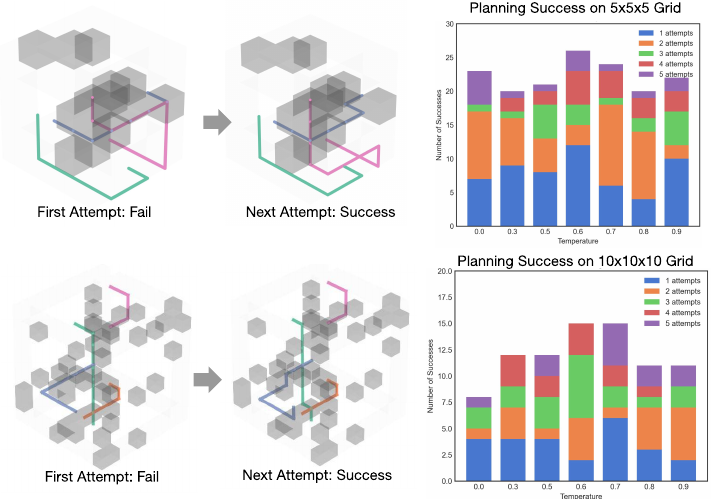}
    \caption{Top right: planning success rate over 30 evaluation runs on a size $5^3$ 3D grid. The LLM (GPT-4)'s performance shows variation with the temperature parameter. Top left: an example output on a size $5^3$ grid. GPT-4 is able to improve the previous plan based on feedback on collided steps, which is appended to the prompt in 3D coordinate format. Bottom: we repeat the same experiments and show example outputs on a more challenging, size $10^3$ grid, where the planning success rate is lower, and the model requires more re-planning attempts before reaching a successful plan.
    }
    \vspace{-2mm}
    \label{fig:appendix-toy-results}
\end{figure}

We run 30 evaluation runs, each with a randomly sampled grid layout. See the left plot in Figure \ref{fig:appendix-toy-results} for evaluation results. We make the following observations: first, the planning success rate varies moderately with the temperature parameter of GPT-4, which controls how deterministic its output generation is. A moderate temperature around 0.6, i.e. partially deterministic, shows the best success rate, and notably is able to 1-shot succeed the most times. Next, we observe the model's ability to read from the feedback of a previous plan, and eventually succeed after a few more attempts. We also provide a visualization of an example output from one of the runs, where the path from first attempt has collision, and the model is able to correct this plan.

%% file: text/appendix-reasoning-dataset.tex
\vspace{-12mm}
\section{Multi-Agent Representation and Reasoning Dataset}
\label{appendix:reasoning}
\vspace{-1mm}
\subsection{Task: Self-Knowledge Question-Answering} 
\subsubsection{Agent Capability}
57 total questions. Sim. Task: sequential transport

\promptblock{\footnotesize 
- Context (system prompt): \\
7 panels on the table, ordered left to right: panel1,...,panel7. They form a straight assembly line, panel1 is closed to panel2 and farthest from panel7. \\
You are robot Alice in front of panel2. You are collaborating with Bob, Chad to sort cubes into their target panels. The task is NOT done until all three cubes are sorted. \\
At current round: \\
blue\_square is on panel5 \\
pink\_polygon is on panel1 \\
yellow\_trapezoid is on panel3 \\
Your goal is to place blue\_square on panel2, but you can only reach panel1, panel2, panel3: this means you can only pick cubes from these panels, and can only place cubes on these panels.\\
Never forget you are Alice! Never forget you can only reach panel1, panel2, panel3!\\
- Question (user prompt): \\
You are Alice. List all panels that are out of your reach. Think step-by-step. Answer with a list of panel numbers, e.g. [1, 2] means you can't reach panel 1 and 2.\\
- Solution: \\
panels [4,5,6,7]
}

\subsubsection{Memory Retrieval}
44 total questions. Task: Make Sandwich, Sweep Floor

\promptblock{\footnotesize 
- Context (system prompt): \newline
[History] \newline
Round\#0: \newline
[Chat History]
[Chad]: ...
[Dave]:...
[Chad]: ...
...
[Executed Action]... \newline
Round\#1: \newline
......\newline
- Current Round \newline
You are a robot Chad, collaborating with Dave to make a vegetarian\_sandwich [......]
You can see these food items are on your reachable side: 
... \newline
- Question (user prompt)
You are Chad. Based on your [Chat History] with Dave and [Executed Action] from previous rounds in [History], what food items were initially on Dave's side of the table? Only list items that Dave explicitly told you about and Dave actually picked up. Don't list items that you are unsure about. Output the item names as a list. Think step-by-step.  \newline
- Solution:  \newline 
bread\_slice1}

\subsection{Challenge: Effective Communication}
\subsection{Inquiry}
41 multiple-choice questions, using Arrange Cabinet task.

\promptblock{\small
- Context (system prompt): \newline
You are Bob, collaborating with Alice, Chad to pick a mug and a cup out of cabinet, and place them on correct coasters.
Both left and right cabinet doors should be OPENed and held open, while anything inside can be PICKed. You must coordinate to complete the task.\newline
At current round:
left door is closed, right door is closed, mug is inside cabinet; cup is inside cabinet; \newline
Alice's gripper is holding nothing, \newline
Your gripper is holding nothing, \newline
Chad's gripper is holding nothing, \newline
Never forget you are Bob! Never forget you can only reach right door handle! \newline
- Question (user prompt):\newline
You are thinking about picking right door handle. Who and what should you ask to confirm this action? Think step-by-step, then choose exactly one option from below. \newline
[A] tell others about this plan because you are free and right door handle is within your reach. \newline
[B] ask if Alice and Chad can reach right door handle because it's not within your reach. \newline
[C] ask if Alice and Chad can help, because you can reach right door handle, but you are busy and they are free. \newline
[D] all three of you are busy, so it's better to wait until later.\newline
\textcolor{blue}{- Solution: [A]}
}

\subsection{Responsiveness}
96 questions. Binary YES or NO  

\promptblock{\footnotesize 
7 panels on the table, ordered left to right: panel1,...,panel7. They form a straight assembly line, panel1 is closed to panel2 and farthest from panel7.\newline 
You are robot Alice in front of panel2. You are collaborating with Bob, Chad to sort cubes into their target panels. The task is NOT done until all three cubes are sorted.
At current round: \newline
blue\_square is on panel5
pink\_polygon is on panel3
yellow\_trapezoid is on panel3 \newline 
Your goal is to place blue\_square on panel2, but you can only reach panel1, panel2, panel3: this means you can only pick cubes from these panels, and can only place cubes on these panels.
You can help others by moving other cubes that are within your reachable panels.
Never forget you are Alice! Never forget you can only reach panel1, panel2, panel3! \newline 
\newline 
You are Alice. Chad needs to put yellow\_trapezoid on panel6, and asked you for help. How should you respond? Think step-by-step. If you can help, your answer must start with YES and propose a plan to help. If you can't help, must answer NO.  
}

\subsection{Challenge: Adaptation to Unexpected Scenarios}
31 questions. Multiple choice (A, B, C). 

\promptblock{\footnotesize 
You are a robot Chad, collaborating with Dave to make a [vegetarian\_sandwich].
Food items must be stacked following this order: bread\_slice1, tomato, cheese, cucumber, bread\_slice2, where bread\_slice1 must be PUT on cutting\_board. 
You must stay on right side of the table! This means you can only PICK food from right side, and Dave can only PICK from the other side.
Only one robot can PUT at a time, so you must coordinate with Dave.
At the current round:
You can see these food items are on your reachable side:
bread\_slice1: on cutting\_board
cheese: atop tomato
tomato: atop bread\_slice1
cucumber: atop cheese
ham: on your side
beef\_patty: on your side
Your gripper is empty \newline 
You are Chad. Your gripper is not working right now. What should you say to Dave? Select exactly one option from below. You must first output a single option number (e.g. A), then give a very short, one-line reason for why you choose it. \newline
Options:\newline
A: Sorry Dave, we can't complete the task anymore, my gripper is broke.\newline
B: Let's stop. The recipe needs ham but Dave can't reach my side and my gripper is not functioning.\newline
\textcolor{dark-green}{C: Dave, go ahead and finish the sandwich without me, there isn't anything we need on my side anyway.}\newline
}